\documentclass[conference]{IEEEtran}
\IEEEoverridecommandlockouts

\usepackage{amsmath,amsfonts,bm}

\def\eqref#1{equation~\ref{#1}}

\def\1{\bm{1}}

\DeclareMathAlphabet{\mathsfit}{\encodingdefault}{\sfdefault}{m}{sl}
\SetMathAlphabet{\mathsfit}{bold}{\encodingdefault}{\sfdefault}{bx}{n}

\usepackage{amsmath,amssymb,amsfonts}
\usepackage{graphicx}
\usepackage{textcomp}
\usepackage{xcolor}
\usepackage{caption}
\usepackage{anyfontsize}
\usepackage{soul}
\usepackage[square, numbers, sort]{natbib}
\usepackage[table, dvipsnames]{xcolor}
\usepackage[font=small, labelfont=bf]{caption}

\usepackage{graphicx}
\usepackage{wrapfig}
\usepackage{multirow}
\usepackage{epsfig}
\usepackage{amsmath}
\usepackage{amssymb}
\usepackage{subfigure}
\usepackage{marvosym}
\usepackage{color}
\usepackage{threeparttable}
\usepackage{algorithm}
\usepackage{algpseudocode}
\usepackage{setspace}
\usepackage{amsthm}
\usepackage{verbatim}
\usepackage{booktabs}
\usepackage{subcaption}
\usepackage[normalem]{ulem}

\usepackage[draft]{hyperref}

\AtBeginDocument{%
  \let\oldref\ref
  \renewcommand{\ref}[1]{\textcolor{blue}{\oldref{#1}}}

  \let\oldpageref\pageref
  \renewcommand{\pageref}[1]{\textcolor{blue}{\oldpageref{#1}}}

  \renewcommand{\cite}[1]{[\textcolor{magenta}{\citenum{#1}}]}
}

\newcommand{\partitle}[1]{\textbf{#1.}}

\newcommand{\boldres}[1]{{\textbf{\textcolor{red}{#1}}}}

\newcommand{\secondres}[1]{\textcolor{blue}{\uline{#1}}}
\def\BibTeX{{\rm B\kern-.05em{\sc i\kern-.025em b}\kern-.08em
    T\kern-.1667em\lower.7ex\hbox{E}\kern-.125emX}}
    
\begin{document}

\title{Airport Terminal Passenger Queue Forecasting \\for Departure Gates and Security Checkpoints
}

\author{\IEEEauthorblockN{Juhwan Lee$^\dagger$, Seokbin Yoon$^\dagger$, Keumjin Lee$^\dagger$, Hojong Baik$^\dagger$ and Seyeon Jung$^\ddagger$}
\IEEEauthorblockA{$^\dagger$Department of Air Transport, Transportation, and Logistics, Korea Aerospace University, Goyang, South Korea}
\IEEEauthorblockA{$^\ddagger$Airport Industry Technology Research Institute, Incheon International Airport Corporation, Incheon, South Korea}
kokomoty97@gmail.com, \{sierra.bin, keumjin.lee, hbaik\}@kau.ac.kr, jessy@airport.kr}

\maketitle

\begin{abstract}
Accurate passenger queue forecasting in airport terminals is essential for efficient departure operations, as it enables proactive congestion management. However, time-varying passenger demand and heterogeneous facility usage across multiple departure facilities make forecasting challenging. In this work, we propose a passenger queue forecasting framework that learns historical passenger flow patterns from operational data. The proposed model employs a Transformer-based architecture to capture temporal dependencies and inter-facility correlations using past queue length and waiting time at departure gates and security checkpoints, together with passenger throughput at check-in islands. The learned representations are mapped to two facility-specific prediction heads to predict queue length and waiting time at departure gates and security checkpoints. Experimental results demonstrate accurate forecasts up to two hours ahead. The proposed approach offers practical real-time decision support for proactive queue management and staff reallocation in airport terminal operations.
\end{abstract}




\section{Introduction}
\begin{figure}[t!]
    \centering
    \includegraphics[width=\linewidth]{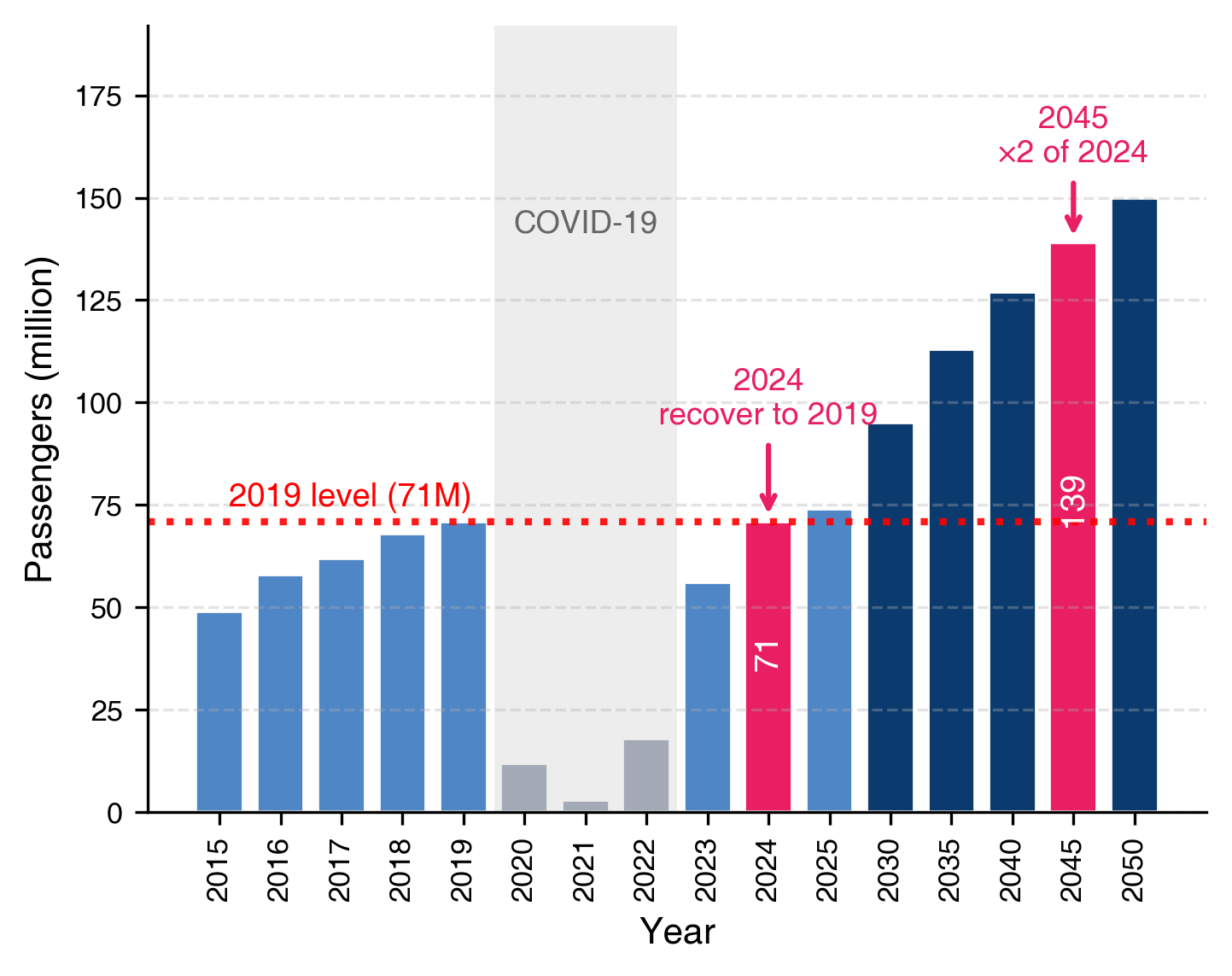}
    \vspace{-7mm}
    \caption{\textbf{Passenger demand forecast for Incheon International Airport}. Passenger volumes dropped drastically during the COVID-19 period but have since rebounded rapidly. Long-term projections indicate continued growth, implying increasing pressure on airport infrastructure and operations.}
    \vspace{-3mm}
    \label{fig:demand_forecast_icn}
\end{figure}

Global air passenger traffic is forecasted to reach 10.2 billion in 2026 and is projected to double by 2045 to reach 18.8 billion passengers~\cite{ACI_WATF_2025_2054}. Consistent with this growth trend, Incheon International Airport (ICN), the largest airport in South Korea, handled 74 million passengers in 2025, surpassing its pre-pandemic record of 71 million in 2019~\cite{IIAC_AviationStats,MOLIT_6thAirportPlan}, as illustrated in Figure~\ref{fig:demand_forecast_icn}. As air traffic demand continues to grow, airports are becoming increasingly congested, leading to longer passenger queues and increased waiting times that can degrade the passenger level of service (LOS)~\cite{RicharddeNeufvilleEtAl2013,Lee2025Dynamic}. Consequently, effective congestion management has become a critical operational challenge~\cite{FelixPatron2021AirportPassengerFlow,MaEtAl2025a}.

A straightforward approach to mitigating airport congestion is to expand terminal infrastructure or increase operational resources to raise capacity~\cite{ALKHEDER2024simmodelingLOS}. Nonetheless, such solutions are limited by physical constraints and typically require substantial cost and long implementation times~\cite{GaterslebenEtAl1999}. In practice, airports must balance operational costs against passenger LOS rather than simply maximizing capacity by operating all available facilities. Consequently, accurately forecasting future congestion is important for supporting resource allocation decisions, such as opening additional facilities or deploying additional staff to maintain acceptable passenger LOS during periods of high demand while avoiding over-provisioning of operational resources. Therefore, congestion forecasting can facilitate timely resource allocation and operational preparedness, helping airports mitigate congestion without the financial and operational burden of capacity expansion~\cite{HOPFE2024102525,Oprea2024discretemodel}.

However, accurately forecasting congestion in large-scale airports remains challenging. Although airports have access to flight schedules and expected passenger volumes in advance, the distribution of passengers across terminal facilities remains highly uncertain. Even when flight schedules and their associated check-in island assignments are known in advance, it remains difficult to determine how passenger flows will propagate through downstream facilities such as security checkpoints and departure gates. Moreover, multiple check-in islands and downstream facilities operate simultaneously under time-varying passenger demand and service capacity, resulting in complex inter-facility interactions whose effects on downstream congestion are difficult to anticipate~\cite{BosscherTowards}. Compounding these difficulties, heterogeneous facility usage patterns and temporal demand fluctuations continuously alter congestion levels across terminal facilities. Such variations can lead to localized congestion at specific facilities even when overall terminal demand remains relatively stable. Collectively, real-time monitoring and congestion management become particularly difficult~\cite{Schultz_PaxDynamic}.

To address such operational complexity, simulation-based models have been widely used in airport terminal operations, particularly discrete-event simulation (DES) and agent-based modeling (ABM), to reproduce passenger movements and evaluate operational scenarios~\cite{ALKHEDER2024simmodelingLOS,GaterslebenEtAl1999,Oprea2024discretemodel,BosscherTowards,Schultz_PaxDynamic,CherednichenkoEtAl2025}. In practice, however, these models often require extensive calibration and rely on user-defined assumptions that are difficult to estimate reliably in practice, such as inter-facility walking times, passenger group sizes, and discretionary facility choices~\cite{Oprea2024discretemodel,THAMPAN2023151}. Thus, these approaches are more suitable for understanding system behavior and conducting post hoc what-if scenario analyses than for proactive queue congestion forecasting.


As an alternative to simulation-based approaches, data-driven models, particularly recurrent neural networks (RNNs) and long short-term memory networks (LSTMs), have recently been applied to airport passenger flow forecasting by learning directly from historical data rather than relying on user-defined simulation assumptions ~\cite{FelixPatron2021AirportPassengerFlow,MaEtAl2025a,HOPFE2024102525}. However, these approaches have primarily focused on predicting aggregate passenger flows, offering limited insight into facility-level congestion dynamics. As a result, they do not explicitly capture interactions among terminal facilities, which restricts their ability to support detailed resource allocation and proactive congestion management in airport terminals. Consequently, existing approaches also remain limited in learning how throughput patterns at upstream processing areas propagate to congestion at downstream facilities. 


To address these challenges, we propose a framework that forecasts passenger queues during the departure process, focusing on two critical congestion points: departure gates and security checkpoints. Specifically, rather than predicting aggregated passenger flows across the terminal, the proposed framework generates facility-level forecasts for each departure gate and security checkpoint. This enables explicit modeling of localized congestion dynamics and heterogeneous facility usage patterns that cannot be captured by aggregate prediction. By leveraging upstream check-in throughput patterns together with historical queue information, the framework is further designed to capture how congestion propagates across terminal facilities. To this end, we utilize the Transformer's self-attention mechanism to capture inter-facility interactions for airport passenger queue forecasting. In particular, the self-attention mechanism models the correlations among the observed queue length and waiting time at departure gates and security checkpoints, as well as the bag-drop throughput at check-in islands. The deep representations learned by the Transformer encoder are then passed to two separate multi-layer perceptrons (MLPs) to forecast future queue dynamics at departure gates and security checkpoints.

The rest of the paper is organized as follows. Section~\ref{sec:Airport_Operations} introduces the airport terminal environment and operational setting considered in this work. Section~\ref{sec:Method} presents the proposed Transformer-based forecasting framework. Section~\ref{sec:Experiments} describes the experimental setup, including the dataset, implementation details, and evaluation metrics, and reports the quantitative results along with further analysis. Finally, Section~\ref{sec:Conclusion} concludes the paper and discusses directions for future work.

\section{Airport Terminal Operations}
\label{sec:Airport_Operations}
This section describes the operational environment and structural characteristics of the airport terminal departure area to motivate the proposed queue forecasting model. The airport terminal is a multi-process system in which passengers sequentially pass through multiple processing facilities, including check-in and baggage processing, passport control, security screening, and immigration~\cite{RicharddeNeufvilleEtAl2013}. Consequently, congestion at a given time cannot be explained solely by demand variations at a single facility; rather, it is shaped by the overall operational conditions of the departure process and the distribution of passenger usage across facilities~\cite{GaterslebenEtAl1999}. To illustrate how such system-level conditions arise in practice, we focus on Incheon International Airport (ICN), South Korea, a large hub airport with complex and time-varying operations.


\begin{figure}[t!]
    \centering
    \includegraphics[width=\linewidth]{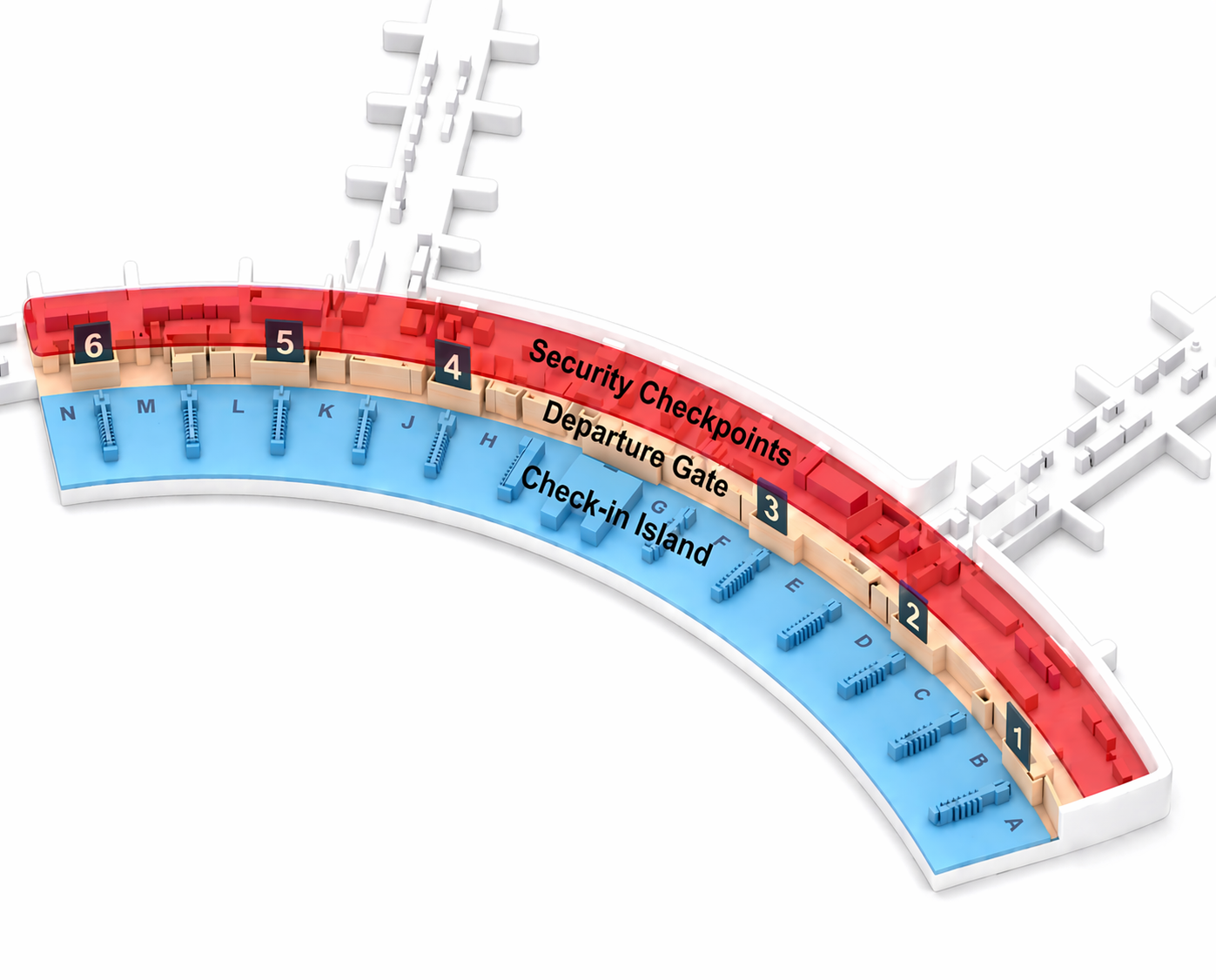}
    \vspace{-7mm}
    \caption{\textbf{Incheon International Airport Terminal 1 layout}. The terminal features multiple check-in islands connected to downstream processing facilities such as departure gates and security checkpoints. This spatial configuration illustrates how passenger flows propagate across facilities within the departure process.}
    \vspace{-3mm}
    \label{fig:fig_ICN}
\end{figure}

\begin{figure*}[t!]
    \centering
    \includegraphics[width=\linewidth]{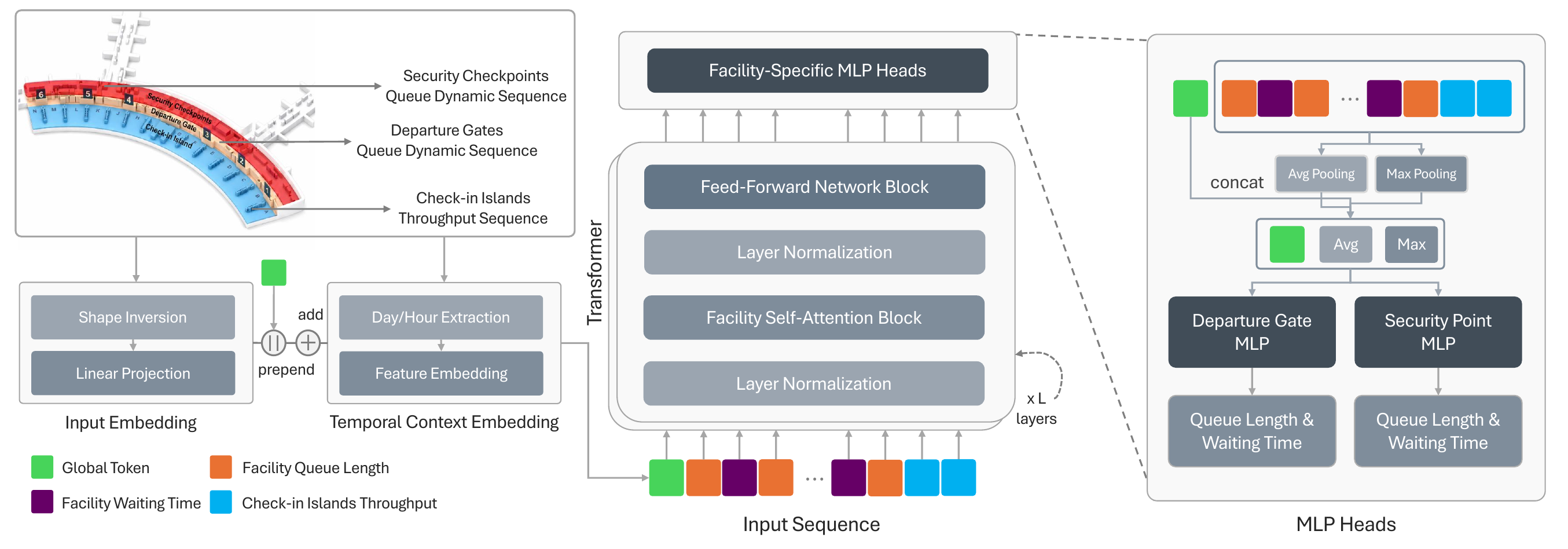}
    \vspace{-7mm}
    \caption{\textbf{Overview of the proposed architecture for airport passenger queue forecasting}. Historical observations from departure gates, security checkpoints, and check-in islands are first embedded into facility-wise tokens, together with a learnable global token. Temporal context features, including day-of-week and hour-of-day information, are then incorporated into the token representations. The resulting input sequence is processed by stacked Transformer encoder layers composed of facility self-attention, layer normalization, and feed-forward network blocks to capture inter-facility dependencies. For prediction, the final global token is concatenated with average-pooled and max-pooled representations of the facility tokens, and the resulting features are passed to two facility-specific MLP heads for forecasting queue length and waiting time at departure gates and security checkpoints.}
    \label{fig:overview}
\end{figure*} 

At ICN, the terminal is characterized by a complex facility layout and operational structure. This study focuses on Terminal 1, which includes 13 check-in islands (A--N, excluding I), each assigned to specific airlines, as well as six departure gates (1--6) and six security checkpoints (1--6), as illustrated in Figure~\ref{fig:fig_ICN}. Because the operating hours of these facilities depend on flight schedules as well as airline and airport policies, passenger demand may become concentrated at specific facilities during certain periods. More importantly, passenger flow across the terminal exhibits strong inter-facility dependencies. The usage patterns of check-in islands affect the downstream distribution of passengers toward departure gates, while the usage patterns of departure gates further influence the distribution of passengers across security checkpoints. Consequently, congestion may arise not only from the overall passenger demand but also from the spatial distribution of passengers across facilities, the interdependencies among these facilities, and the available processing capacity of each facility~\cite{RicharddeNeufvilleEtAl2013}.

Therefore, forecasting future passenger queue dynamics without considering inter-facility correlations is inadequate. A reliable forecasting model should jointly account for time-varying terminal operations and heterogeneous passenger flows across facilities~\cite{AnagnostopoulouEtAl2024}. In particular, changes in the usage patterns of multiple check-in islands can induce time-varying inflow patterns to downstream facilities, namely departure gates and security checkpoints, while the historical queue states of these downstream facilities also provide important information for future congestion. Motivated by this operational setting, we aim to forecast future queue length and waiting time at departure gates and security checkpoints by modeling inter-facility relationships within the complex sequential processes of the airport terminal. Such forecasts can support more effective staff allocation and proactive congestion mitigation in terminal operations.

\section{Methodology}
\label{sec:Method}
We first formulate the passenger queue forecasting problem, briefly review the Transformer's self-attention mechanism, and then present our neural network architecture for passenger queue forecasting for departure gates and security checkpoints. An overview of the proposed architecture is illustrated in Figure~\ref{fig:overview}.

\subsection{Problem Formulation}
In airport passenger queue forecasting, the input sequence is constructed by combining historical observations from departure gates, security checkpoints, and check-in islands. Given past observations over $T$ time steps, at each time step $t$, we concatenate queue-related features and upstream passenger flow information into a unified representation $\mathbf{x}_t = [\mathbf{x}_t^{\mathrm{DG/SC}},\mathbf{x}_t^{\mathrm{CI}}]$, $\mathbf{x}_t^{\mathrm{DG/SC}} \in \mathbb{R}^{2F}$ contains queue length and waiting time for all service facilities, with $F = M + N$ denoting the total number of departure gates and security checkpoints, and $\mathbf{x}_t^{\mathrm{CI}} \in \mathbb{R}^{C}$ denotes passenger throughput at $C$ check-in islands. The final input sequence is then defined as 
$\mathbf{X} = \{\mathbf{x}_1, \ldots, \mathbf{x}_T\} \in \mathbb{R}^{T \times (2F + C)}$. The forecasting objective is to predict future queue length and waiting time separately for departure gates and security checkpoints over the next $S$ time steps:
$\mathbf{Y}_{\mathrm{DG}} \in \mathbb{R}^{S \times (2M)}$ and 
$\mathbf{Y}_{\mathrm{SC}} \in \mathbb{R}^{S \times (2N)}$.  We train a forecaster \(f_{\theta}\), parameterized by learnable parameters \(\theta\), to model the relationship between historical observations \(\mathbf{X}\) and future targets:
\begin{equation}
\widehat{\mathbf{Y}}_{\mathrm{DG}},\ \widehat{\mathbf{Y}}_{\mathrm{SC}} = f_{\theta}(\mathbf{X}).
\end{equation}
The model is trained by solving the following optimization problem:
\begin{equation}
\theta^* = \arg\min_{\theta} \ \mathcal{L}\big(\mathbf{Y}_{\mathrm{DG}}, \widehat{\mathbf{Y}}_{\mathrm{DG}}\big) + 
\mathcal{L}\big(\mathbf{Y}_{\mathrm{SC}}, \widehat{\mathbf{Y}}_{\mathrm{SC}}\big),
\end{equation}
where $\mathcal{L}(\cdot, \cdot)$ denotes a loss function that measures the discrepancy between the ground-truth targets and the corresponding predictions.

\subsection{Self-Attention Mechanism}
The core operation of the Transformer is the self-attention mechanism (SA) based on scaled dot-product operations, which enables the model to capture dependencies within the data and learn informative representations~\cite{vaswani2017attention}. The SA is defined as follows:
\begin{equation}
    \mathrm{Attention}(\mathbf{Q}, \mathbf{K}, \mathbf{V}) = \mathrm{Softmax}\left(\frac{\mathbf{Q}\mathbf{K}^{\top}}{\sqrt{d_k}}\right)\mathbf{V},
\end{equation}
where $\mathbf{Q}$, $\mathbf{K}$, and $\mathbf{V}$ denote the query, key, and value matrices, respectively, obtained by linear projections of the input sequence $\mathbf{H} \in \mathbb{R}^{T \times D}$, i.e., $\mathbf{Q} = \mathbf{H}\mathbf{W}^Q$, $\mathbf{K} = \mathbf{H}\mathbf{W}^K$, and $\mathbf{V} = \mathbf{H}\mathbf{W}^V$, where $\mathbf{W}^Q, \mathbf{W}^K, \mathbf{W}^V \in \mathbb{R}^{D \times D}$ are learnable parameter matrices. Here, $T$ denotes the sequence length, $D$ is the model's latent dimension, and $d_k$ is the key dimension used to scale the dot-product attention for numerical stability. To enhance the model's representation capacity, we employ multi-head SA (MSA), which allows the model to jointly attend to information from different representation subspaces. Specifically, the input tokens are split across $H$ heads along the latent dimension $D$, each head performs SA independently on a lower-dimensional subspace, and the outputs are then concatenated and projected back to the original dimension:
\begin{equation}
\mathrm{MultiHead}(\mathbf{Q}, \mathbf{K}, \mathbf{V}) = \mathrm{Concat}(\mathrm{head}_1, \ldots, \mathrm{head}_H)\mathbf{W}^O,
\end{equation}
where $\mathrm{head}_h = \mathrm{Attention}(\mathbf{Q}\mathbf{W}_h^Q, \mathbf{K}\mathbf{W}_h^K, \mathbf{V}\mathbf{W}_h^V)$ and $\mathbf{W}^O \in \mathbb{R}^{D \times D}$. This enables the model to capture diverse relationships across different subspaces.

\subsection{Proposed Neural Network Architecture}
In this work, we propose a novel Transformer-based architecture that explicitly models interactions among check-in islands for airport departure queue forecasting at departure gates and security checkpoints. The proposed model mainly consists of an input embedding, a Transformer encoder, and two separate prediction heads for departure gates and security checkpoints, respectively. We describe the core components in detail below.

\partitle{Input Embedding} Given the historical queue dynamics and passenger throughput sequence $\mathbf{X} \in \mathbb{R}^{T \times (2F +C)}$, where $P = 2F + C$ hereafter, which is a multivariate time series serving as the input, we first invert the sequence and then project it into a higher-dimensional space through a linear layer before passing it to the model. That is, we transpose $\mathbf{X}$ to $\mathbf{X}^{\top} \in \mathbb{R}^{P \times T}$ and embed it into $\mathbf{H}^0 = \{\mathbf{h}_1^0, \ldots, \mathbf{h}_P^0\} \in \mathbb{R}^{P \times D}$, resulting in $P$ variate tokens, each with dimensionality $D$, following the Inverted Transformer (iTransformer)~\cite{liu2023itransformer}, as follows:
\begin{equation}
    \mathbf{H}^0 = \mathbf{X}^{\top}\mathbf{W} + \mathbf{b},
\end{equation}
where $\mathbf{W} \in \mathbb{R}^{T \times D}$ and $\mathbf{b} \in \mathbb{R}^{D}$ are embedding weights and bias vectors. This design contrasts with most Transformer-based forecasters, which treat observations across multiple variates at the same timestamp as a single temporal token. Aggregating multiple variates into a single temporal token may introduce irrelevant information, hindering pattern extraction. Furthermore, such representations may have limited expressiveness in capturing long-term dependencies due to their localized receptive field, even with MSA.

Following this design, rather than relying solely on variate tokens for forecasting, we prepend a global token $\mathbf{h}_0 \in \mathbb{R}^{D}$ to the variate sequence $\mathbf{H}^0$, yielding $\mathbf{H}^0 = \{\mathbf{h}_0^0, \mathbf{h}_1^0, \ldots, \mathbf{h}_P^0\} \in \mathbb{R}^{ (P+1) \times D}$. This choice is motivated by the role of the global token as a learnable representation that aggregates information across all tokens via the subsequent Transformer encoder’s SA~\cite{devlin2019bert, dosovitskiy2020image}. 

Finally, we incorporate temporal context by embedding calendar features such as day-of-week and hour-of-day using learnable embedding layers. Let $d$ and $h$ denote the day-of-week and hour-of-day indices, respectively. We obtain their embeddings as
\begin{equation}
\mathbf{e}^{(d)} = \mathrm{Emb}_d(d), \quad \mathbf{e}^{(h)} = \mathrm{Emb}_h(h),
\end{equation}
where $\mathrm{Emb}_d$ and $\mathrm{Emb}_h$ are embedding functions. The combined temporal embedding is then given by $\mathbf{e}^{(t)} = \mathbf{e}^{(d)} + \mathbf{e}^{(h)}$. We add this to each token in $\mathbf{H}^0$ in an element-wise manner:
\begin{equation}
\mathbf{h}_i \leftarrow \mathbf{h}_i + \mathbf{e}^{(t)}, \quad \forall i \in \{0, \ldots, P\}.
\end{equation}
This enables the model to capture periodic patterns and time-dependent variations in passenger queue dynamics that are not explicitly encoded in the input sequence.

\begin{figure*}[t!]
    \centering
    \includegraphics[width=\linewidth]{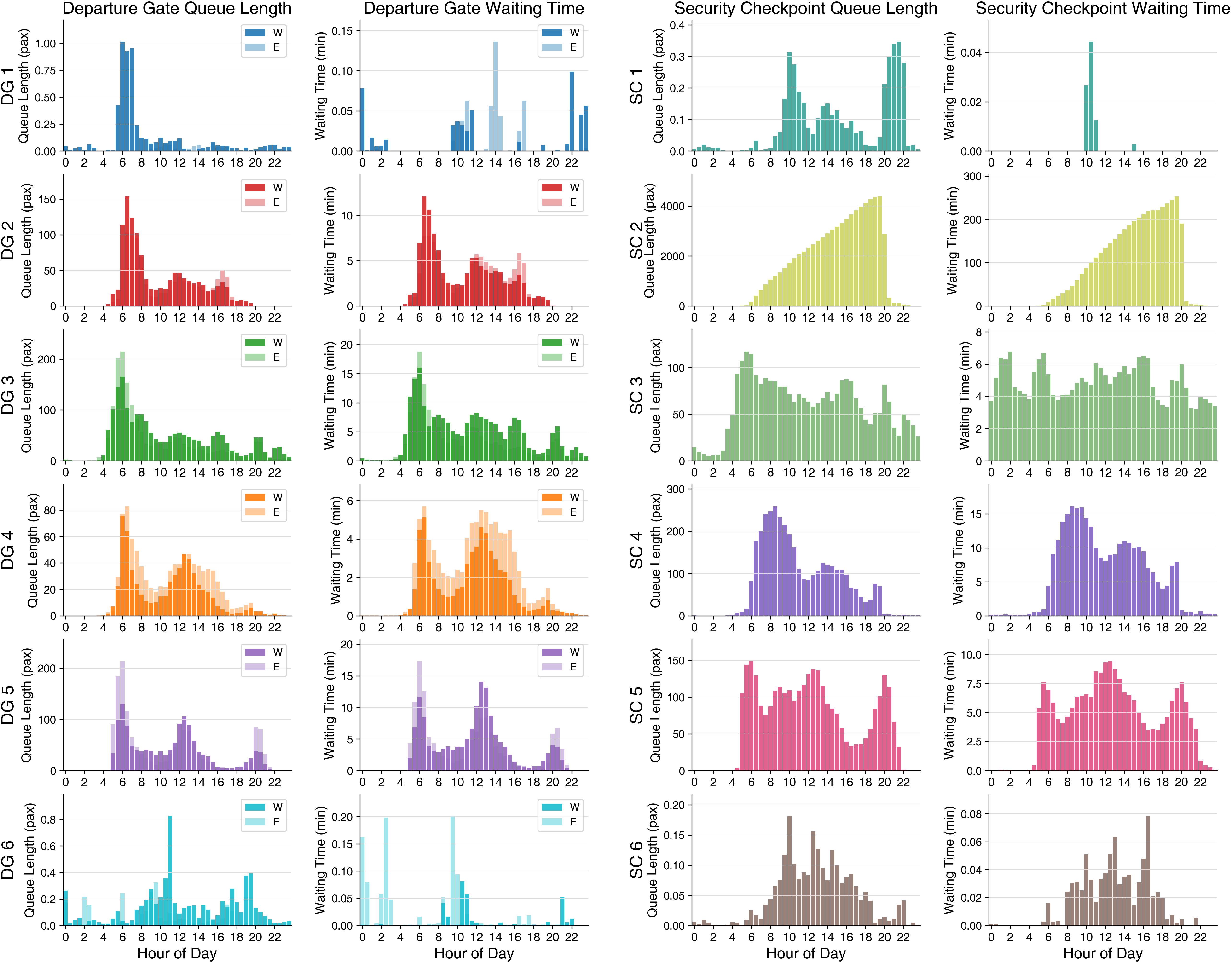}
    \vspace{-5mm}
    \caption{\textbf{Overview of the passenger flow and queue data}. Eight months of statistics for queue length (pax) and waiting time (minutes) are shown for each departure gate (first two columns) and security checkpoint (last two columns) at ICN.}
    \label{fig:statistics}
\end{figure*}

\partitle{Transformer Encoder} After input embedding, we pass the embedded variate sequence $\mathbf{H}^0$ into $L$ identical Transformer encoder layers, each consisting of layer normalization (LN)~\cite{ba2016layer}, MSA, and a feed-forward network (FFN) in sequence. The LN is first applied to the representation of each variate token to improve training stability and convergence as follows:
\begin{equation}
\mathrm{LayerNorm}(\mathbf{H}^{\ell}) = 
\left\{
\frac{\mathbf{h}_p^{\ell} - \mathrm{Mean}(\mathbf{h}_p^{\ell})}
{\sqrt{\mathrm{Var}(\mathbf{h}_p^{\ell})}}
\;\middle|\; p = 0, \ldots, P
\right\},
\end{equation}
where $\mathrm{Mean}(\mathbf{h}_p^{\ell})$ and $\mathrm{Var}(\mathbf{h}_p^{\ell})$ are computed over the latent dimension of each token at the $\ell$-th layer. The Facility MSA models the interactions among the $P+1$ variate tokens, each corresponding to a service facility. Consequently, facilities with stronger correlations receive higher attention weights, allowing their information to be more effectively incorporated into each token representation. Finally, the FFN is applied independently to each variate token to capture nonlinear representations. It consists of two linear transformations with a nonlinear activation function such as Gaussian error linear units (GELU)~\cite{hendrycks2016gaussian}, expanding and projecting the model dimension as $\mathbb{R}^{D} \rightarrow \mathbb{R}^{4D} \rightarrow \mathbb{R}^{D}$ as follows:
\begin{equation}
\mathrm{FFN}(\mathbf{H}^{\ell}) = \mathrm{GELU}(\mathbf{H}^{\ell} \mathbf{W}^{\mathrm{in}} + \mathbf{b}^{\mathrm{in}})\mathbf{W}^{\mathrm{out}} + \mathbf{b}^{\mathrm{out}},
\end{equation}
where $\mathbf{W}^{\mathrm{in}} \in \mathbb{R}^{D \times 4D}$, $\mathbf{W}^{\mathrm{out}} \in \mathbb{R}^{4D \times D}$, $\mathbf{b}^{\mathrm{in}} \in \mathbb{R}^{4D}$, and $\mathbf{b}^{\mathrm{out}} \in \mathbb{R}^{D}$. The GELU activation is defined as $\mathrm{GELU}(x) = x \cdot \Phi(x)$, where $\Phi(\cdot)$ denotes the cumulative distribution function of the standard normal distribution.


\partitle{Prediction Heads} After passing the sequence of variate tokens through the $L$ Transformer encoder layers, we obtain the final representation $\mathbf{H}^L$. We then extract the global token $\mathbf{h}_0^{L}$ from $\mathbf{H}^L$ and use it as input to two separate MLPs, corresponding to departure gates and security checkpoints. We additionally compute max pooling and average pooling over the variate tokens $\{\mathbf{h}_1^L, \mathbf{h}_2^L, \ldots, \mathbf{h}_P^L\}$, excluding the global token. Since the global token aggregates information only via SA, this complementary pooling provides more direct summary statistics. Specifically, we obtain $\mathbf{h}_{\mathrm{avg}}^L \in \mathbb{R}^{D}$ and $\mathbf{h}_{\max}^L \in \mathbb{R}^{D}$, which denote the element-wise average and maximum over the $P$ variate tokens, respectively.

We then concatenate the three representations and feed them into each MLP:
\begin{equation}
\widehat{\mathbf{Y}}_{\mathrm{DG}} = \mathrm{MLP}_{\mathrm{DG}}\Big([\mathbf{h}_0^L,\ \mathbf{h}_{\mathrm{avg}}^L,\ \mathbf{h}_{\max}^L]\Big),
\end{equation}
\begin{equation}
\widehat{\mathbf{Y}}_{\mathrm{SC}} = \mathrm{MLP}_{\mathrm{SC}}\Big([\mathbf{h}_0^L,\ \mathbf{h}_{\mathrm{avg}}^L,\ \mathbf{h}_{\max}^L]\Big).
\end{equation}
Each MLP consists of multiple linear layers interspersed with GELU activations, and outputs $\widehat{\mathbf{Y}}_{\mathrm{DG}} \in \mathbb{R}^{S \times (2M)}$ and $\widehat{\mathbf{Y}}_{\mathrm{SC}} \in \mathbb{R}^{S \times (2N)}$ over the next $S$-step prediction horizon for $M$ departure gates and $N$ security checkpoints. Since both queue length and waiting time are predicted, the output dimensions are $2M$ and $2N$, respectively.

\section{Experiments}
\label{sec:Experiments}
\subsection{Experimental Setup} 
\partitle{Dataset} We use passenger flow and queue data collected at ICN, from January to August 2025. The dataset is acquired from the passenger flow management system (PFMS), and a video-based monitoring system (XOVIS). PFMS provides time-stamped records of passenger activities, including boarding pass issuance and bag-drop transactions. XOVIS supplies facility-level measurements of queue length and waiting time at 10-minute intervals across terminal facilities, using computer vision sensors to continuously estimate congestion states. In total, the dataset covers baggage drop transactions at 13 check-in islands, queue measurements at 6 departure gates (each divided into west and east lines, resulting in 12 gate-level queues), and queue measurements at 6 security checkpoints within the terminal. An overview of the dataset is shown in Figure~\ref{fig:statistics}, which presents statistics of queue length and waiting time across departure gates and security checkpoints.


\begin{table*}[t]
\centering
\begin{minipage}{0.48\textwidth}
  \centering
  \caption{Forecasting results for \textbf{departure gate queue lengths}. Results are averaged from all prediction lengths.}
  \label{tab:dg_queue}

  \vspace{-1mm}
  \renewcommand{\arraystretch}{1.2}
  \setlength{\tabcolsep}{3.5pt}
  \setlength{\extrarowheight}{1pt}
  \begin{small}
  \resizebox{0.98\linewidth}{!}{

  \begin{tabular}{c|cc|cc|cc|cc}
    \toprule
    Models & \multicolumn{2}{c}{\textbf{Ours}} & \multicolumn{2}{c}{FFN} & \multicolumn{2}{c}{Seq2Seq LSTM} & \multicolumn{2}{c}{Transformer} \\
    \cmidrule(lr){2-3}\cmidrule(lr){4-5}\cmidrule(lr){6-7}\cmidrule(lr){8-9}
    Metric & MAE & RMSE & MAE & RMSE & MAE & RMSE & MAE & RMSE \\
    \toprule
    DG2-W & \boldres{6.176} & \boldres{10.979} & 8.826 & 16.213 & 7.879 & 14.435 & \secondres{6.247} & \secondres{11.606} \\
    DG2-E & \boldres{2.362} & \boldres{6.143} & 3.264 & 8.844 & 2.846 & 7.807 & \secondres{2.732} & \secondres{7.015} \\
    DG3-W & \boldres{8.708} & \boldres{13.504} & 13.906 & 20.678 & 12.579 & 19.723 & \secondres{9.364} & \secondres{15.533} \\
    DG3-E & \boldres{6.299} & \boldres{10.947} & 9.555 & 17.835 & 8.806 & 16.233 & \secondres{6.899} & \secondres{13.201} \\
    DG4-W & \secondres{4.309} & \boldres{8.157} & 5.799 & 11.864 & 5.090 & 9.989 & \boldres{4.207} & \secondres{8.239} \\
    DG4-E & \secondres{5.820} & \boldres{9.664} & 8.760 & 14.205 & 7.865 & 13.418 & \boldres{5.737} & \secondres{9.774} \\
    DG5-W & \boldres{8.619} & \boldres{13.603} & 12.919 & 20.364 & 11.970 & 19.400 & \secondres{9.035} & \secondres{14.831} \\
    DG5-E & \boldres{5.598} & \boldres{11.225} & 9.438 & 19.350 & 8.274 & 17.791 & \secondres{6.539} & \secondres{13.367} \\
    \midrule
    Avg & \boldres{5.986} & \boldres{10.528} & 9.058 & 16.169 & 8.164 & 14.850 & \secondres{6.345} & \secondres{11.696} \\
    \bottomrule
  \end{tabular}
  }
  \end{small}
\end{minipage}
\hfill
\begin{minipage}{0.48\textwidth}
  \centering
  \caption{Forecasting results for \textbf{departure gate waiting time}. Results are averaged from all prediction lengths.}
  \label{tab:dg_wait}

  \vspace{-1mm}
  \renewcommand{\arraystretch}{1.2}
  \setlength{\tabcolsep}{3.5pt}
  \setlength{\extrarowheight}{1pt}
  \begin{small}
  \resizebox{0.98\linewidth}{!}{

  \begin{tabular}{c|cc|cc|cc|cc}
    \toprule
    Models & \multicolumn{2}{c}{\textbf{Ours}} & \multicolumn{2}{c}{FFN} & \multicolumn{2}{c}{Seq2Seq LSTM} & \multicolumn{2}{c}{Transformer} \\
    \cmidrule(lr){2-3}\cmidrule(lr){4-5}\cmidrule(lr){6-7}\cmidrule(lr){8-9}
    Metric & MAE & RMSE & MAE & RMSE & MAE & RMSE & MAE & RMSE \\
    \toprule
    DG2-W & \boldres{0.817} & \boldres{1.669} & 1.180 & 2.694 & 1.052 & 2.208 & \secondres{0.877} & \secondres{2.133} \\
    DG2-E & \boldres{0.344} & \secondres{0.915} & 0.457 & 1.189 & 0.398 & 1.079 & \secondres{0.360} & \boldres{0.869} \\
    DG3-W & \boldres{0.945} & \boldres{1.531} & 1.457 & 2.311 & 1.304 & 2.169 & \secondres{0.962} & \secondres{1.684} \\
    DG3-E & \boldres{0.720} & \boldres{1.215} & 1.024 & 1.857 & 0.954 & 1.691 & \secondres{0.736} & \secondres{1.377} \\
    DG4-W & \secondres{0.479} & \secondres{0.880} & 0.599 & 1.115 & 0.580 & 1.019 & \boldres{0.447} & \boldres{0.790} \\
    DG4-E & \secondres{0.620} & \secondres{1.069} & 0.934 & 1.569 & 0.829 & 1.509 & \boldres{0.605} & \boldres{1.025} \\
    DG5-W & \boldres{0.883} & \boldres{1.383} & 1.304 & 2.092 & 1.216 & 1.977 & \secondres{0.929} & \secondres{1.479} \\
    DG5-E & \boldres{0.627} & \boldres{1.168} & 0.892 & 1.699 & 0.802 & 1.564 & \secondres{0.653} & \secondres{1.248} \\
    \midrule
    Avg & \boldres{0.679} & \boldres{1.229} & 0.981 & 1.816 & 0.892 & 1.652 & \secondres{0.696} & \secondres{1.326} \\
    \bottomrule
  \end{tabular}
  }
  \end{small}
\end{minipage}
\end{table*}

\begin{table*}[t]
\centering
\begin{minipage}[t]{0.48\textwidth}
    \vspace{0pt}

  \centering
  \caption{Forecasting results for \textbf{security checkpoint queue lengths}. Results are averaged from all prediction lengths.}
  \label{tab:sc_queue}
  \vspace{-1mm}
  \renewcommand{\arraystretch}{1.2}
  \setlength{\tabcolsep}{3.5pt}
  \setlength{\extrarowheight}{1pt}
  \begin{small}
  \resizebox{0.98\linewidth}{!}{
  \begin{tabular}{c|cc|cc|cc|cc}
    \toprule
    Models & \multicolumn{2}{c}{\textbf{Ours}} & \multicolumn{2}{c}{FFN} & \multicolumn{2}{c}{Seq2Seq LSTM} & \multicolumn{2}{c}{Transformer} \\
    \cmidrule(lr){2-3}\cmidrule(lr){4-5}\cmidrule(lr){6-7}\cmidrule(lr){8-9}
    Metric & MAE & RMSE & MAE & RMSE & MAE & RMSE & MAE & RMSE \\
    \toprule
    SC3 & \boldres{9.358} & \boldres{12.438} & 15.629 & 19.719 & 14.859 & 19.320 & \secondres{9.884} & \secondres{14.026} \\
    SC4 & \boldres{7.654} & \boldres{11.448} & 13.728 & 18.906 & 12.607 & 18.149 & \secondres{8.154} & \secondres{12.453} \\
    SC5 & \boldres{9.710} & \boldres{13.918} & 18.107 & 24.506 & 17.009 & 24.064 & \secondres{10.967} & \secondres{16.443} \\
    \midrule
    Avg & \boldres{8.907} & \boldres{12.601} & 15.821 & 21.044 & 14.825 & 20.511 & \secondres{9.668} & \secondres{14.307} \\
    \bottomrule
  \end{tabular}
  }
  \end{small}
\end{minipage}
\hfill
\begin{minipage}[t]{0.48\textwidth}
\vspace{0pt}

  \centering
  \caption{Forecasting results for \textbf{security checkpoint waiting time}. Results are averaged from all prediction lengths.}
  \label{tab:sc_wait}
  \vspace{-1mm}
  \renewcommand{\arraystretch}{1.2}
  \setlength{\tabcolsep}{3.5pt}
  \setlength{\extrarowheight}{1pt}
  \begin{small}
  \resizebox{0.95\linewidth}{!}{

  \begin{tabular}{c|cc|cc|cc|cc}
    \toprule
    Models & \multicolumn{2}{c}{\textbf{Ours}} & \multicolumn{2}{c}{FFN} & \multicolumn{2}{c}{Seq2Seq LSTM} & \multicolumn{2}{c}{Transformer} \\
    \cmidrule(lr){2-3}\cmidrule(lr){4-5}\cmidrule(lr){6-7}\cmidrule(lr){8-9}
    Metric & MAE & RMSE & MAE & RMSE & MAE & RMSE & MAE & RMSE \\
    \toprule
    SC3 & \secondres{0.699} & \secondres{0.929} & 1.047 & 1.391 & 1.001 & 1.310 & \boldres{0.673} & \boldres{0.923} \\
    SC4 & \secondres{0.668} & \boldres{1.166} & 1.065 & 1.716 & 0.995 & 1.652 & \boldres{0.639} & \secondres{1.308} \\
    SC5 & \boldres{0.683} & \boldres{0.922} & 1.090 & 1.452 & 1.042 & 1.420 & \secondres{0.710} & \secondres{0.975} \\
    \midrule
    Avg & \secondres{0.683} & \boldres{1.006} & 1.067 & 1.520 & 1.013 & 1.461 & \boldres{0.674} & \secondres{1.069} \\
    \bottomrule
  \end{tabular}
  }
  \end{small}
\end{minipage}
\end{table*}

\partitle{Preprocessing} For preprocessing, we first exclude data from departure gate numbers 1 and 6 (for both west and east lines) and from security checkpoints 1 and 6, as these facilities are designated for transportation-vulnerable passengers and exhibit queue dynamics that differ systematically from regular passenger flows\footnote{The current dataset does not provide passenger-level labels identifying transportation-vulnerable travelers, which makes it difficult to model their distinct queue behaviors reliably within the present framework. We plan to incorporate these facilities into the analysis in future work.}. Furthermore, security checkpoint 2 is excluded from the analysis because queue measurements obtained from XOVIS contain substantial noise and frequently exhibit anomalous observations and implausible values, which we acknowledge as a limitation of the current dataset. Therefore, the number of departure gates, security checkpoints, and check-in islands considered in the analysis are $M=8$, $N=3$, and $C=13$, respectively. We then apply domain-informed filtering rules to the remaining data to remove implausible or corrupted observations. Finally, we restrict the data to operational hours between 05:00 and 22:00 KST (local time) to exclude late-night and early-morning periods, during which passenger volume is minimal and only a limited subset of departure gates is in operation. 

We then construct input–output pairs, where the input $\mathbf{X}$ consists of past queue length and waiting time at departure gates and security checkpoints, together with passenger throughput at check-in islands, over a three-hour window ($T=18$), and the target $\mathbf{Y}$ corresponds to future queue length and waiting time over a two-hour prediction horizon ($S=12$) for both departure gates and security checkpoints. The dataset is subsequently split into training, validation, and test sets using an 8:1:1 ratio. 


\partitle{Implementation Details}
We set the number of encoder layers to $L=3$, the latent dimension to $D=256$, and the number of attention heads to $H=4$. Two MLP decoders for departure gates and security checkpoints, respectively, each composed of three fully connected layers with GELU activations applied between successive layers for nonlinearity. During training, the batch size is set to 32, and we use the AdamW optimizer~\cite{kingma2014adam} with a learning rate of $1\times10^{-4}$ for 300 epochs, minimizing the mean squared error (MSE) between the ground-truth sequence and the predicted sequence.

\partitle{Baseline Comparisons}
We compare our approach with three representative deep learning-based forecasting models, including a FFN, a Seq2Seq LSTM~\cite{sutskever2014sequence}, and an encoder–decoder Transformer~\cite{vaswani2017attention}, which model temporal dependencies using (i) fully connected, (ii) recurrent, and (iii) self-attention mechanisms, respectively. All baseline models are configured under the same experimental settings as the proposed approach to ensure a fair comparison. To evaluate the forecasting performance, we use the mean absolute error (MAE) and root mean squared error (RMSE). 


\begin{figure*}[t!]
    \centering
    \includegraphics[width=\linewidth]{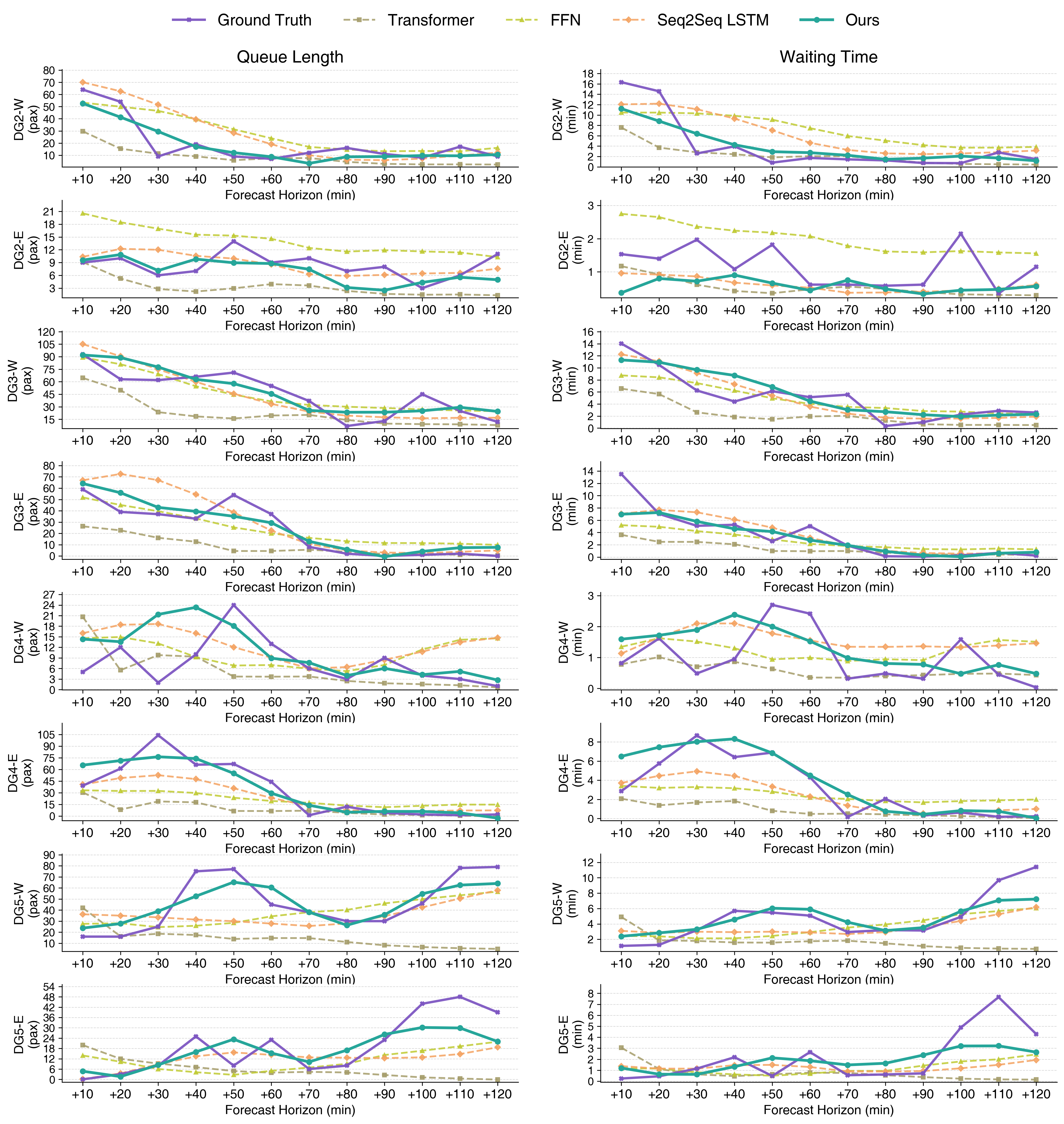}
    \vspace{-5mm}
    \caption{Visualization of the forecasting results for \textbf{queue length and waiting time at departure gates DG2--DG5 for both the west and east lines}. The left panel shows the queue length, while the right panel shows the waiting time.}
    \label{fig:pred1}
\end{figure*} 

\begin{figure*}[t!]
    \centering
    \includegraphics[width=\linewidth]{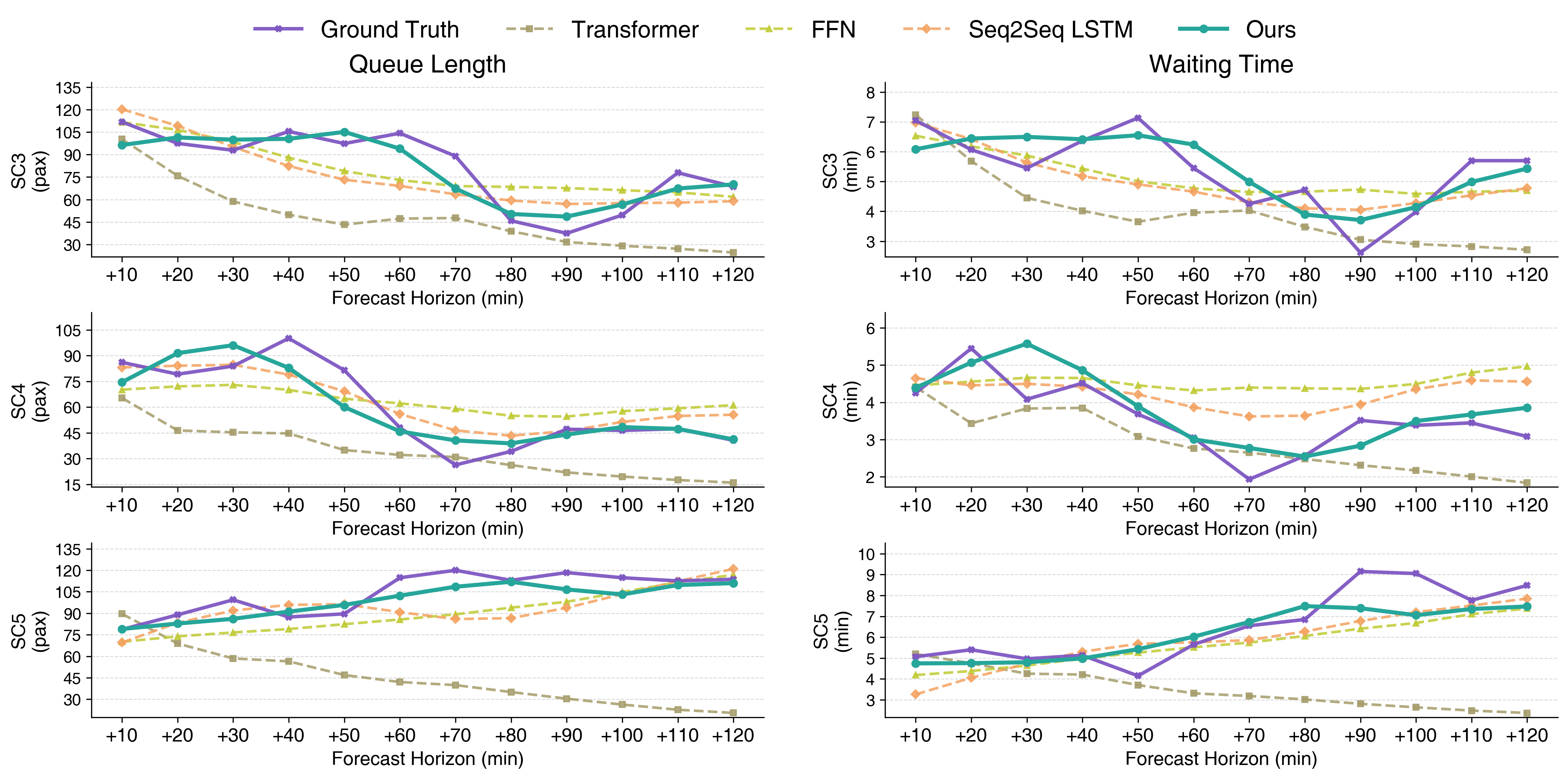}
    \vspace{-5mm}
    \caption{Visualization of the forecasting results for \textbf{queue length and waiting time at security checkpoints SC3--SC5}. The left panel shows the queue length, while the right panel shows the waiting time.}
    \label{fig:pred2}
\end{figure*} 
\subsection{Quantitative Results}
To better analyze forecasting performance, we report the results for queue length and waiting time at each departure gate and security checkpoint, as summarized in Tables~\ref{tab:dg_queue}--\ref{tab:sc_wait}. Here, queue length is measured in passengers (pax) and waiting time in minutes. We can observe that our model achieves highly competitive performance across all queue length and waiting time forecasting tasks at both departure gates and security checkpoints compared to FFN, Seq2Seq LSTM, and Transformer. In general, the proposed model consistently yields lower errors across most facilities, indicating its robustness in capturing diverse passenger flow patterns.

Particularly, our approach significantly outperforms FFN and Seq2Seq LSTM, demonstrating the effectiveness of the attention-based architecture for passenger queue forecasting. For example, in the DG3-W queue length forecasting case (Table~\ref{tab:dg_queue}), which exhibits the highest errors among all departure gates across all models, our model achieves $8.708$ pax (MAE) and $13.504$ pax (RMSE). This corresponds to improvements of $\underline{37.4}$\% and $\underline{34.7}$\% over FFN, and $\underline{30.8}$\% and $\underline{31.5}$\% over Seq2Seq LSTM, in terms of MAE and RMSE, respectively. We attribute this improvement to the limited ability of these baseline models to selectively process the input sequence, which may lead them to focus on irrelevant or less influential service facilities.

Compared to Transformer, which achieves the second-best performance overall, our model consistently better RMSE results and generally better MAE results. The SA in Transformer enables the model to selectively focus on important time steps for prediction and thus yields competitive performance, but passenger queue forecasting is influenced more strongly by facility usage patterns, namely which facilities are in operation and how they are connected, than by temporal dependencies alone. This suggests that explicitly modeling inter-facility relationships is more effective than relying only on temporal attention over the input sequence. These advantages are reflected in the average RMSE results across all four tasks. Our model achieves $10.528$ pax, $1.229$ minutes, $12.601$ pax, and $1.006$ minutes for departure gate queue length, departure gate waiting time, security checkpoint queue length, and security checkpoint waiting time, respectively, corresponding to average RMSE improvements of $\underline{10.0}\%$, $\underline{7.3}\%$, $\underline{11.9}\%$, and $\underline{5.9}\%$ over Transformer. Similar improvements are also observed in MAE for three of the four tasks, with comparable performance on security checkpoint waiting time.

\subsection{Qualitative Results}
We next showcase the forecasting results to compare the predictive behaviors of different models. Figures~\ref{fig:pred1} and~\ref{fig:pred2} provide representative prediction examples from the proposed model, FFN, Seq2Seq LSTM, and Transformer. As shown in the figures, the proposed model more accurately captures future temporal variations and consistently outperforms the competing baselines. Figure~\ref{fig:pred1} presents the forecasting results for queue length and waiting time at departure gates (DG2--DG5) for both west and east lines). Compared with the baseline models, our model more accurately captures the overall trends of future queue dynamics. Transformer tends to underestimate the target values in many departure-gate cases, whereas the FFN and Seq2Seq LSTM baselines often show limited ability to capture local variations, particularly when the target series exhibits noticeable temporal changes. This behavior is particularly evident in highly volatile cases such as DG4, where the target series exhibits sharp rises and drops. Overall, these results indicate that the proposed model is more effective at preserving abrupt temporal changes in highly variable series.

Figure~\ref{fig:pred2} presents the corresponding forecasting results for security checkpoints (SC3--SC5). Similar tendencies are observed in both queue length and waiting time predictions. The proposed model more accurately reproduces the decreasing trends, stabilization phases, and moderate recoveries in the ground truth, whereas the baseline models show a more limited ability to capture these patterns. These qualitative observations are consistent with the quantitative results and indicate that the proposed model more effectively captures the temporal evolution of passenger congestion across different facilities. Although the current model still fails to perfectly capture all future dynamics, it follows the overall temporal trends more faithfully than the baseline models, and further architectural improvements may help close the remaining gap.

\begin{figure*}[t!]
    \centering
    \includegraphics[width=\linewidth]{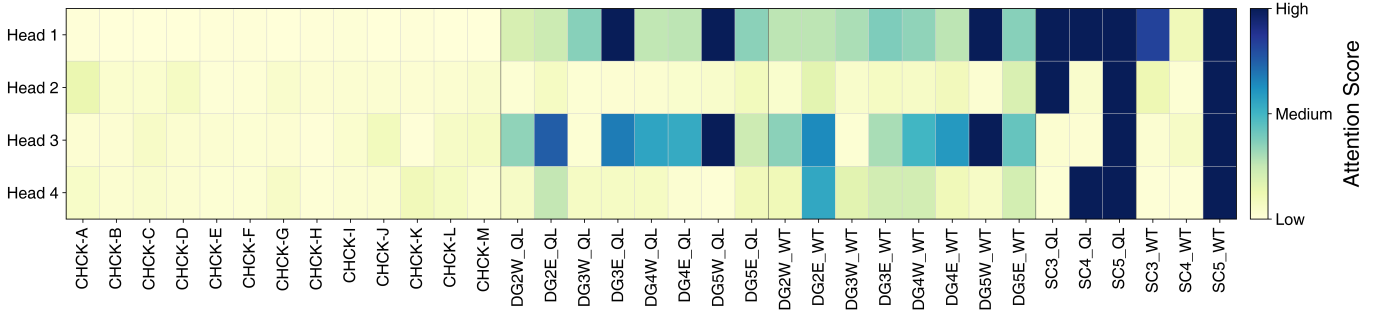}
    \vspace{-5mm}
    \caption{\textbf{Head-specific global token attention patterns}. The heatmap shows the average attention weights from the global token to facility tokens across the test set. Different attention heads exhibit distinct attention preferences toward departure gates and security checkpoints.}
    \label{fig:attention}
\end{figure*} 

\begin{figure}[t!]
    \centering
    \includegraphics[width=\linewidth]{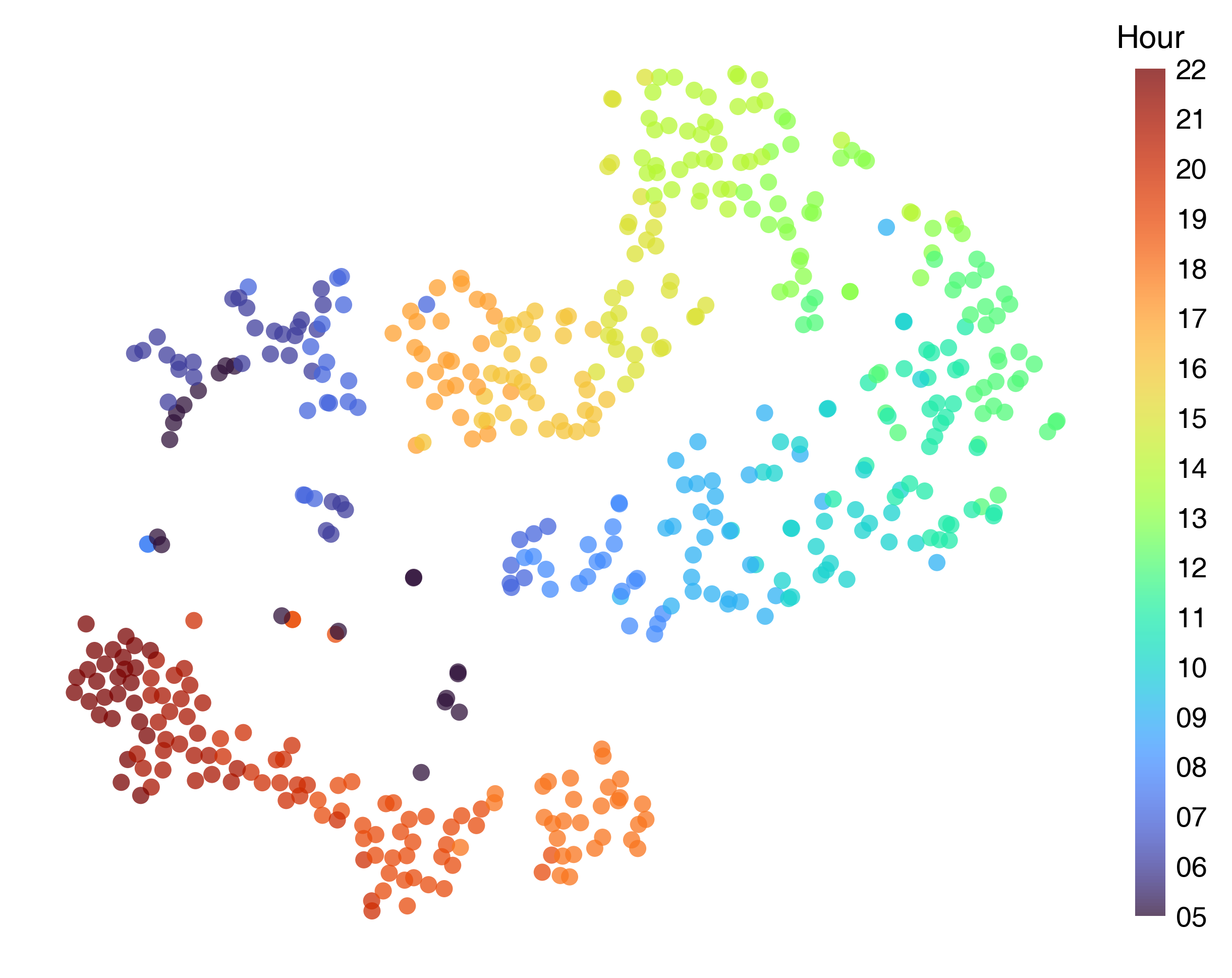}
    \vspace{-7mm}
    \caption{\textbf{t-SNE visualization of the learned representations}. Each point corresponds to a test sample and is colored according to its time-of-day label.} 
    \vspace{-3mm}
    \label{fig:fig_tsne}
\end{figure} 

\subsection{Representation Analysis}
To better understand the internal mechanisms of the proposed model, we analyze the attention distributions of the global token across different attention heads. Specifically, we examine how the global token attends to different facility tokens, including check-in islands, departure gates, and security checkpoints. Figure~\ref{fig:attention} shows the global token's attention scores assigned by the global token to facility tokens. As shown in the figure, Head 1 attends broadly to both departure gate and security checkpoint tokens, whereas Heads 2 and 4 place greater emphasis on security checkpoint tokens. In contrast, Head 3 focuses primarily on departure gate queue length and waiting time tokens. These observations suggest that different attention heads capture complementary aspects of passenger congestion dynamics. Although check-in islands receive comparatively less attention, this behavior is expected because the forecasting targets correspond to departure gate and security checkpoint congestion, while check-in islands mainly provide auxiliary information.

Furthermore, we evaluate the quality of the learned representations. Specifically, we pass all test samples through the encoder, extract the final-layer representation vectors, and project them into a two-dimensional space using t-distributed stochastic neighbor embedding (t-SNE)~\cite{van2008visualizing}. To investigate whether temporal characteristics are faithfully encoded in the learned representation space, the projected samples are colored according to their time-of-day labels. Figure~\ref{fig:fig_tsne} reveals a clear temporal structure, where samples associated with similar times of day tend to cluster together, whereas temporally distant samples are generally separated in the embedding space. For example, samples corresponding to 14:00 and 22:00 are located far apart, reflecting their different operational conditions. Interestingly, although 06:00 and 22:00 are separated by a similar temporal gap, their representations are located much closer together. This observation is consistent with airport operations, as both periods correspond to relatively low-demand conditions near the start or end of daily operations. These results suggest that the proposed model learns meaningful representations that capture operationally relevant temporal patterns in passenger congestion dynamics.

\section{Conclusion}
\label{sec:Conclusion}
We present a data-driven framework for airport passenger queue forecasting. The proposed approach learns passenger flow dynamics from historical operational data and forecasts future queue lengths and waiting times at departure gates and security checkpoints by modeling cross-facility dependencies and upstream passenger flow patterns. The framework captures interactions among terminal facilities and heterogeneous passenger flow patterns, enabling accurate forecasts of future congestion conditions under complex airport operating environments. We evaluate the proposed framework using real-world passenger flow and queue data collected from the passenger flow management system and a video-based monitoring system at Incheon International Airport. Experimental results demonstrate that the proposed method can accurately forecast passenger queue conditions up to two hours ahead, consistently outperforming feed-forward, recurrent, and Transformer-based forecasting baselines. 

Our empirical findings underscore the importance of modeling inter-facility relationships within airport terminals. Consistent performance gains across all forecasting tasks suggest that accounting for interactions among check-in islands, departure gates, and security checkpoints provides a more informative representation of passenger flow dynamics. These findings indicate that passenger congestion is not solely driven by temporal demand variations but is also strongly affected by interactions among terminal facilities. Consequently, incorporating such inter-facility dependencies can significantly enhance forecasting accuracy.

Several compelling directions exist to extend the scope and operational utility of this framework. First, incorporating explicit contextual features such as real-time flight schedules and dynamic facility operating rates would likely improve forecasting performance during sudden demand shifts. Second, we aim to explore alternative deep learning architectures, such as graph neural networks, to more precisely encode the physical topology and spatial constraints of terminal layouts. Furthermore, investigating the robustness and adaptability of the model under irregular operations, such as severe flight delays, sudden cancellations, or unexpected facility closures, remains a critical next step. Future research should also examine the generalizability of the proposed framework across airports with different terminal layouts and operational environments. Finally, integrating queue forecasts into downstream decision-support systems for resource allocation, staff scheduling, and congestion mitigation may enable more proactive and efficient airport terminal operations.

\section*{Acknowledgment}
This work is supported by the Korea Agency for Infrastructure Technology Advancement (KAIA) grant funded by the Ministry of Land, Infrastructure and Transport (Grant RS-2022-00156364).

\hypersetup{linkcolor=Red, urlcolor=Blue}
\bibliographystyle{ieeetr}
\small{
\bibliography{ref}
}
\end{document}